\documentclass[10pt, a4paper]{article}

\usepackage{lrec-coling2024}
\usepackage{inconsolata}

\usepackage{amsmath}
\usepackage{mathtools}

\usepackage{graphicx}
\usepackage{multirow}
\usepackage{amsfonts}

\usepackage{xcolor}
\definecolor{cyan}{RGB}{121, 180, 160}

\title{Auxiliary Knowledge-Induced Learning for Automatic Multi-Label Medical Document Classification}

\name{Xindi Wang$^{1,2}$, Robert E. Mercer$^{1}$, Frank Rudzicz$^{2,3,4}$} 

\address{$^1$Department of Computer Science, University of Western Ontario, Canada\\
$^2$ Vector Institute for Artificial Intelligence, Canada\\
$^3$ Faculty of Computer Science, Dalhousie University, Canada\\
$^4$ Department of Computer Science, University of Toronto, Canada\\
         xwang842@uwo.ca, mercer@csd.uwo.ca, frank@dal.ca}

\abstract{
The International Classification of Diseases (ICD) is an authoritative medical classification system of different diseases and conditions for clinical and management purposes. ICD indexing assigns a subset of ICD codes to a medical record. Since human coding is labour-intensive and error-prone, many studies employ machine learning to automate the coding process. ICD coding is a challenging task, as it needs to assign multiple codes to each medical document from an extremely large hierarchically organized collection. In this paper, we propose a novel approach for ICD indexing that adopts three ideas: (1) we use a multi-level deep dilated residual convolution encoder to aggregate the information from the clinical notes and learn document representations across different lengths of the texts; (2) we formalize the task of ICD classification with auxiliary knowledge of the medical records, which incorporates not only the clinical texts but also different clinical code terminologies and drug prescriptions for better inferring the ICD codes; and (3) we introduce a graph convolutional network to leverage the co-occurrence patterns among ICD codes, aiming to enhance the quality of label representations. Experimental results show the proposed method achieves state-of-the-art performance on a number of measures.
\\ \newline  }

\begin{document}

\maketitleabstract

\section{Introduction}
Electronic health records (EHRs)\footnote{\urlstyle{same}\url{https://www.cms.gov/Medicare/E-Health/EHealthRecords}} contain all of the key administrative clinical data relevant to a person's care under a particular provider, including demographics, past history notes, progress notes, laboratory reports, diagnoses, and medications. EHRs have been increasingly used in a variety of settings which provide opportunities to enhance patient care and facilitate clinical research. The International Classification of Diseases  (ICD)\footnote{\urlstyle{same}\url{https://www.who.int/standards/classifications/classification-of-diseases}} is often used as a surrogate for  clinical outcomes of interest, as it is designed to provide diagnostic assistance and classify  health disorders. ICD is a medical classification taxonomy maintained by the World Health Organization (WHO)\footnote{\urlstyle{same}\url{https://www.who.int}}, which serves a broad range of uses in diagnostic processes, epidemiology, health management, and other clinical activities. %The first published version of ICD is in 1893, and it has become one of the most important indexing systems in medical management and healthcare related research. 
There are two types of codes in the ICD coding system, namely procedure codes\footnote{\urlstyle{same}\url{https://en.wikipedia.org/wiki/Procedure_code}} (that are used to identify  specific surgical, medical, or diagnostic interventions) and diagnosis codes\footnote{\urlstyle{same}\url{https://en.wikipedia.org/wiki/Diagnosis_code}} (that are used to identify diseases, disorders and symptoms). In the 10$^{\textrm{th}}$ edition, there are over 70,000 procedure codes and over 69,000 diagnosis codes\footnote{\urlstyle{same}\url{https://www.cdc.gov/nchs/icd/icd10cm_pcs.htm}}.%, and ICD codes are revised periodically.

The task of ICD indexing aims to associate ICD codes with EHR documents. Currently, ICD indexing is carried out manually by human annotators, which is labour-intensive and error-prone \cite{b0be0de6987549188fdfae53aa2a6554}. Therefore, automatic annotation has gained interest in the research community. Automatic ICD indexing can be regarded as an extreme multi-label text classification (XMTC) problem, where each EHR document can be labeled with multiple ICD codes. Compared with standard multi-label classification tasks, XMTC finds relevant labels  from an extremely large set of labels. Large-scale ICD indexing is severely challenged by several problems. First, the distribution of ICD codes is extremely long-tailed: while some ICD codes occur frequently, many others seldom appear, if at all, because of the rarity of the diseases. For instance, among the 942 unique 3-digit ICD codes in the MIMIC-III \citelanguageresource{Johnson2016MIMICIIIAF} dataset (the largest publicly available medical dataset), the ten most common codes account for 26\% of all code occurrences and the least common 437 codes account for only 1\% \cite{Bai2019ImprovingMC}. Second, unstructured clinical texts are long (containing an average of 1596 words in the MIMIC-III dataset) and noisy (including irrelevant information, misspellings, and non-standard abbreviations). These  difficulties make extracting relevant information from clinical texts, for all ICD codes, very challenging.  

\begin{figure}[t]
\includegraphics[width=\columnwidth]{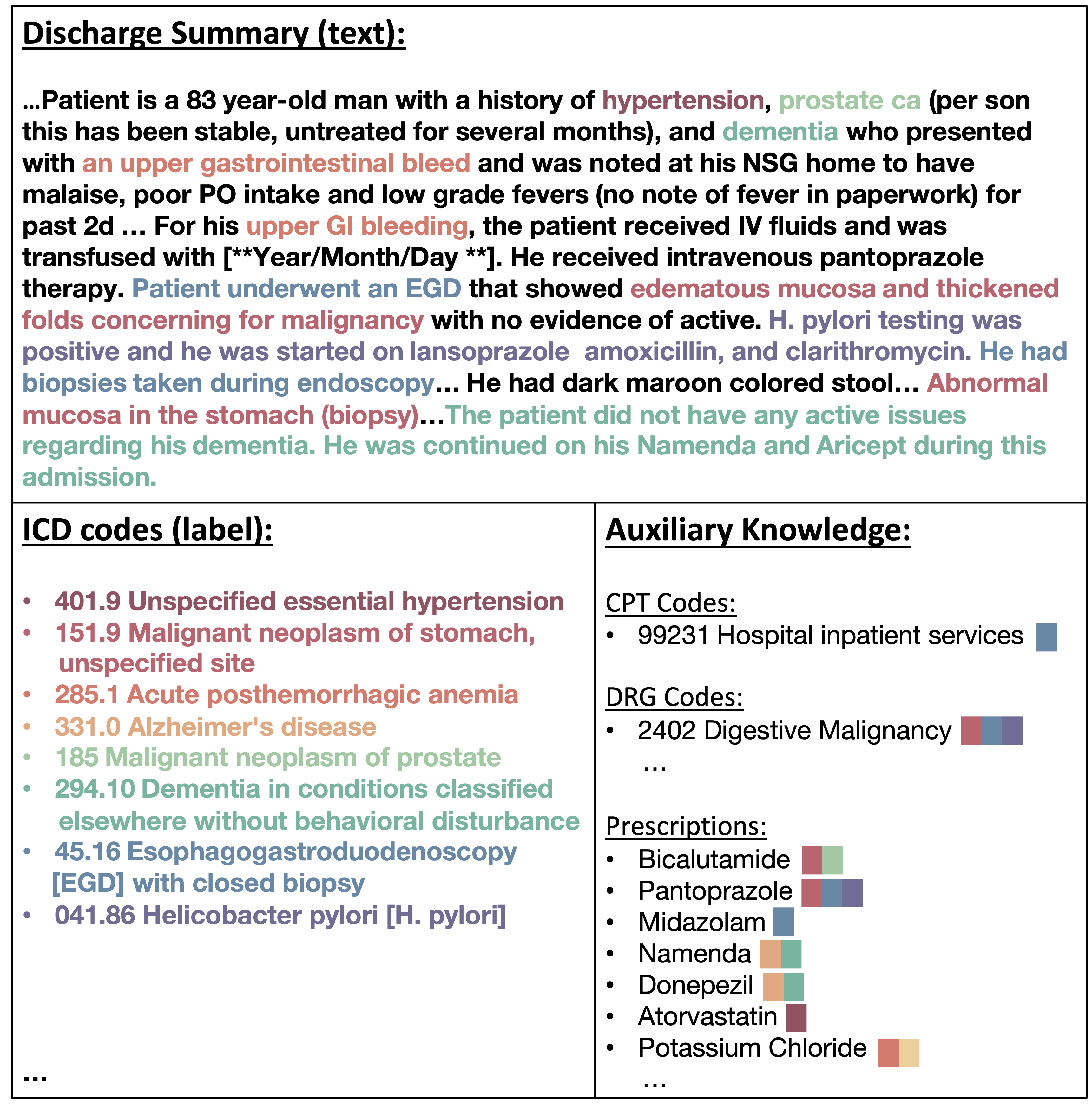}
\caption{An example of a patient record from the MIMIC-III dataset which includes the discharge summary, assigned ICD codes and auxiliary knowledge. We colour each code and its corresponding mentions in the discharge summary and auxiliary knowledge. We use the auxiliary knowledge of the notes to predict relevant codes of summary.}
\label{fig:1}
\end{figure}

We propose a novel auxiliary knowledge-induced medical code labelling architecture to address these issues. To lessen the problems caused by the long-tailed distribution of ICD codes, we leverage code co-occurrence and join auxiliary knowledge with the clinical texts %, which can be leveraged 
to improve coding accuracy. 

\textbf{Code Co-occurrence} The co-occurrence of codes in clinical texts provides valuable insights into the relationships between different diseases or conditions. For instance  Figure \ref{fig:1} shows that the code for ``Dementia in conditions classified elsewhere without behavioral disturbance'' (294.10) can be easily captured from the text (i.e., the highlighted words in \textcolor{cyan}{desaturated cyan}). However, inferring the code for  ``Alzheimer's disease'' (331.0) is more challenging as the clues are less explicit. Fortunately, there is a strong association between these two diseases, with ``Alzheimer's disease'' being one of the most common causes of ``dementia''. This association can be captured by leveraging the fact that the codes for these two diseases often co-occur in clinical texts. By leveraging code co-occurrence patterns, we can capture the dependencies and correlations among codes. This allows us to better understand the context in which specific codes occur and make more accurate predictions based on these relationships beyond using only the clinical texts themselves.

\textbf{Auxiliary Knowledge} EHR auxiliary knowledge is widely available, but is often overlooked in previous studies. In addition to clinical texts, an EHR document is also associated with various auxiliary knowledge such as code systems (other than ICD codes) and drug prescriptions. Specifically, we are interested in two code terminologies (diagnosis-related group (DRG)\footnote{\urlstyle{same}\url{https://www.cms.gov/Medicare/Medicare-Fee-for-Service-Payment/AcuteInpatientPPS/MS-DRG-Classifications-and-Software}} codes and current procedural terminology (CPT)\footnote{\urlstyle{same}\url{https://www.ama-assn.org/amaone/cpt-current-procedural-terminology}} codes), as well as the medications prescribed to patients, which could be strong indicators of ICD predictions. For instance, Figure \ref{fig:1} shows ``Namenda'' in drug prescriptions, which would strongly suggest the patient is most likely to have Alzheimer's disease. By incorporating auxiliary knowledge, we augment the information available for coding tasks. This external knowledge provides additional context and insights that can aid in accurately assigning  appropriate ICD codes.

To alleviate the long text issue, we introduce a multi-level dilated residual convolutional network to ensure the extracted representations focus on the long clinical notes. With multi-level dilation rates, convolutions can capture broader contexts while preserving spatial resolution, enabling the network to have a larger receptive field without increasing the number of parameters. 

The contributions of this paper are as follows:
\begin{itemize}
\setlength{\itemsep}{-2pt}
    \item We propose a framework that is capable of simultaneously dealing with both long-tail and long-text issues in the ICD prediction task. 
    \item To alleviate the long-tail issue, we propose a graph convolutional network to leverage  code co-occurrence which captures the %internal 
    connections among codes with different frequencies. We %further 
    integrate external knowledge %to build 
    using an auxiliary knowledge mask which constrains the large space of possible ICD codes. 
    \item To handle long texts, we use a multi-level dilated residual convolutional network, enabling the model to capture long-range dependencies and local context with different dilation rates.
    \item We evaluate on a widely used automatic ICD coding dataset, MIMIC-III, and the results show that our proposed model outperforms previous methods on a number of measures.
\end{itemize}

\section{Related Work}
\subsection{Automatic ICD Indexing}
Automatic ICD indexing is a long-standing task in the healthcare domain. To the best of our knowledge, the earliest work was proposed by \citet{Larkey1996CombiningCI}, who combined three classifiers (K-nearest-neighbour, relevance feedback, and Bayesian independence classifiers) to automatically assign ICD codes to dictated inpatient discharge summaries. \citet{10.1145/288627.288649} proposed a hierarchical model that used the topology of the code structure, and then calculated the cosine similarity of TF-IDF representations between clinical texts and ICD codes. A variety of rule-based methods \cite{10.5555/1572392.1572416,Farkas2008AutomaticCO} and statistical machine learning algorithms, such as support vector machines \cite{lita-etal-2008-large}, were later applied to the ICD coding task. 

With deep neural networks, many previous works have proven the effectiveness of convolutional neural networks (CNNs), recurrent neural networks (RNNs), and their variants for ICD coding. \citet{mullenbach-etal-2018-explainable} combined a CNN with an attention mechanism to capture relevant information in the clinical texts for each ICD code. \citet{10.1145/3357384.3357897} further improved the CNN attention model by incorporating multi-scale feature attention. Many other CNN variants were proposed to deal with lengthy and noisy clinical texts, such as MultiResCNN \cite{Li2020ICDCF}, DCAN \cite{ji-etal-2020-dilated}, and EffectiveCAN \cite{liu-etal-2021-effective}. MultiResCNN introduced a multi-filter residual CNN to capture text patterns of different lengths and used  a residual convolutional layer to enlarge the receptive field. DCAN stands for `dilated convolutional attention network', which used a single filter and the dilation operation to control the receptive field. EffectiveCAN used a CNN based encoder with squeeze-and-excitation networks together with residual networks to aggregate the information across clinical texts. RNN-based models, which have also been widely used in the ICD coding task, are able to capture contextual information across input texts. \citet{Shi2017TowardsAI} proposed a character-aware long short-term memory (LSTM) recurrent network to learn the representations of the clinical texts. \citet{xie-xing-2018-neural} used a tree-of-sequences LSTM architecture and adversarial learning to capture hierarchical relationships among ICD codes. \citet{Baumel2018MultiLabelCO} presented a hierarchical attention-bidirectional gated recurrent unit (HA-GRU) to label a document by identifying the sentences relevant for each ICD code. LAAT \cite{ijcai2020-461} used a bidirectional Long-Short Term Memory (BiLSTM) encoder and customized label-wise attention mechanism to learn label-specific vectors across  clinical text fragments. 

To tackle the hierarchical relationships among  ICD codes, graph convolutional neural networks (GCNNs) \cite{Kipf:2016tc} can be employed. For instance, \citet{rios-kavuluru-2018-shot} and \citet{10.1145/3357384.3357897} leveraged GCNN to capture both the hierarchical relationships among ICD codes and the semantics of each code. HyperCore \cite{Cao2020HyperCoreHA} considered both code hierarchy and code co-occurrence to learn code representations in the co-graph by exploiting the GCNN.  Our work does not consider the ICD hierarchy because the parent-child hierarchy is shallow, i.e., the ICD hierarchy has only three levels. Also, the possibility of a parent code and a child code both being assigned to the same discharge summary is essentially zero in the MIMIC-III dataset, since the child code is a more specific description of the parent code.

Besides employing ICD code information, some other external knowledge has also been considered. For instance, \citet{Bai2019ImprovingMC} proposed a knowledge source integration (KSI) model that incorporates external knowledge from Wikipedia to calculate matching scores between a clinical note and disease-related Wikipedia documents, in order to obtain useful information for ICD predictions. \citet{yuan-etal-2022-code} proposed a multiple synonym matching network (MSMN) to leverage synonyms of the ICD codes for better code representation learning. \citet{yang-etal-2022-knowledge-injected} further incorporated a pre-trained language model with three domain-specific knowledge sources: code hierarchy, synonyms, and abbreviations to help the code classification.

\subsection{Extreme Multi-label Text Classification}
Extreme Multi-label Text Classification (XMTC) is designed to assign relevant labels to objects from an extremely large set of potential labels. Deep learning methods have been employed for XMTC tasks to learn semantic representations of text. For instance, XML-CNN \cite{10.1145/3077136.3080834} used a 1-dimensional convolutional network with different vertical filters and dynamic pooling to learn the text representations. The model utilizes convolutional filters of varying sizes (vertical filters), which operate across different n-gram sizes. Dynamic pooling is then added to allow the model to handle input texts of varying lengths by aggregating the feature maps produced by the convolutional layers into fixed-size representations.  Furthermore, AttentionXML \cite{10.5555/3454287.3454810} employs a BiLSTM layer followed by an attention mechanism, a strategy designed to capture the most relevant text features for each label. \citet{wang-etal-2022-kenmesh} introduced a knowledge-enhanced mask attention module in the KenMeSH framework, designed to refine the candidate label set by reducing its size. This innovative module leverages external knowledge to guide the attention mechanism, focusing it on the most relevant labels for a given text. By filtering out less pertinent labels, the model can concentrate on a more manageable subset of candidates, %significantly improving the accuracy and efficiency of predictions. 
effectively improving the accuracy of the predictions. 

\begin{figure*}[t]
\begin{center}
\includegraphics[width=\textwidth]{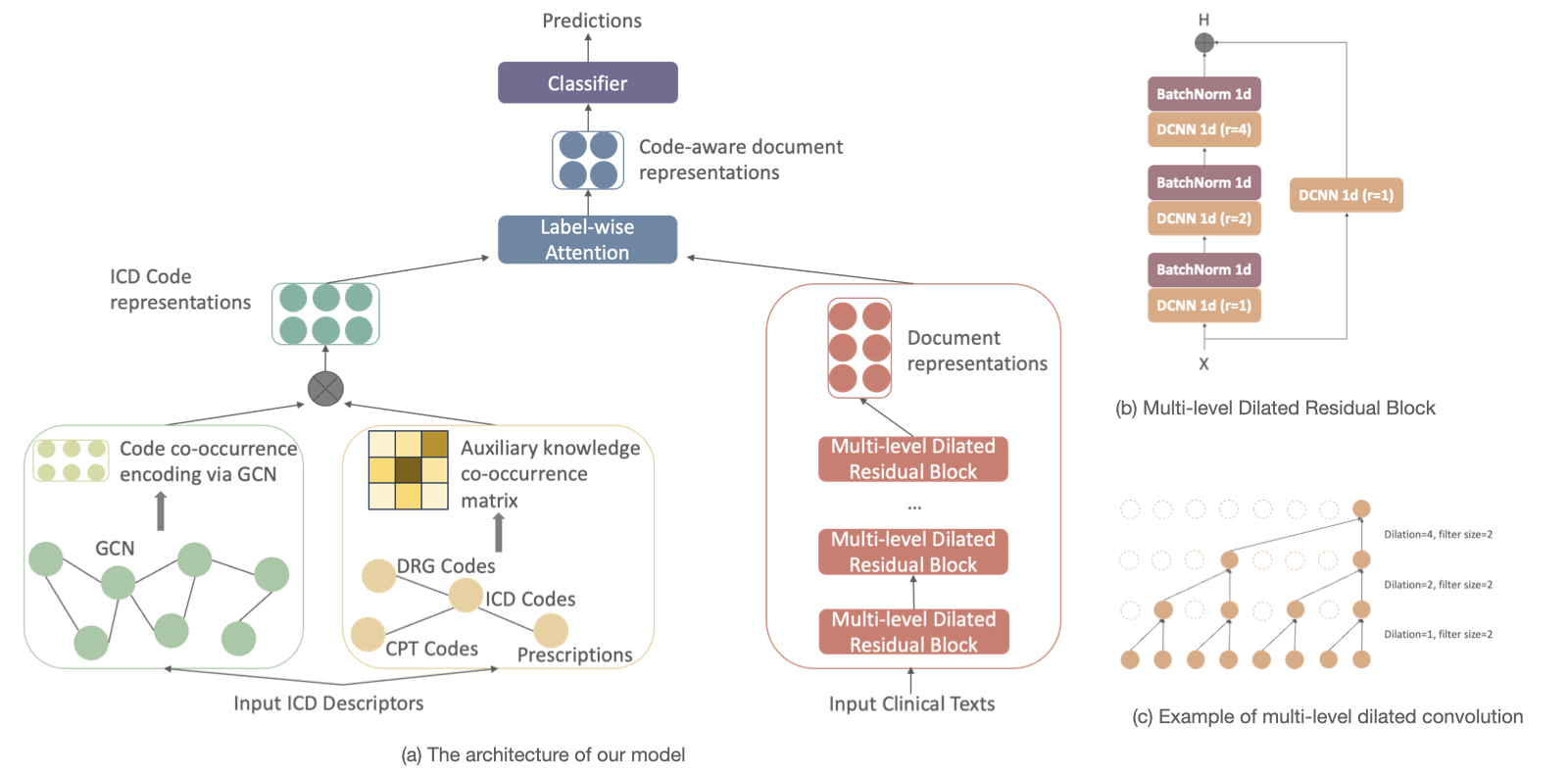}
\caption{The architecture of our model. There are four main components in our method: a document encoder that contains multiple multi-level dilated residual blocks, a label encoder that includes label co-occurrence representation learned by GCN and an auxiliary knowledge mask, a label-wise attention layer and a classifier.}
\label{fig:2}
\end{center}
\end{figure*}

\section{Method}

We treat ICD indexing as an extreme multi-label text classification problem in which a set of medical records $\mathcal{X}= \{x_{1}, x_{2}, ..., x_{N}\}$ and a set of ICD codes $\mathcal{Y} = \{y_{1}, y_{2}, ..., y_{L}\}$ is given. The objective of multi-label classification is to learn $L$ binary classifiers in which each classifier is to determine $y_j \in \{0, 1\}$ using the training set $\mathcal{D} = \{(x_{i}, Y_{i})\}$, $Y_{i} \subset \mathcal{Y}$, $i = 1, ..., N$, where $j$ is the $j$-th label in $\mathcal{Y}$, and $N$ is the number of records in the set. 

In this section, we present a neural architecture for ICD indexing shown in Figure \ref{fig:2}(a). Our model is composed of a clinical text encoder that extracts the long-term dependencies and generates higher-level semantic representations for each clinical text, a label encoder that utilizes label co-occurrence relations and auxiliary knowledge to generate dynamic code presentations for each clinical note, a label-wise attention layer that produces the code-aware document representation, and a classifier that produces the final predictions of the ICD codes. 

\subsection{Clinical Text Encoder}

\subsubsection{Input Layer} 
Our model leverages a clinical record $C$ as the input that consists of a sequence of words $\{w_1, w_2, ..., w_n\}$, where $n$ is the sequence length. The embedding matrix $ \tilde{E} \in \mathbb{R}^{d_e}$ is pre-trained using word2vec \cite{10.5555/2999792.2999959} from the raw texts of the dataset, where $d_{e}$ is the dimension of the word vectors. A word $w_i$ in the medical record corresponds to an embedding vector $e_i$ by looking up $\tilde{E}$. Therefore, the word embedding matrix for the input medical record is $E = \{e_1, e_2, ..., e_n\} \in \mathbb{R}^{n \times {d_e}}$, where $n$ is the sequence length.

\subsubsection{Multi-level Dilated Residual Block} 
To transform the clinical record into informative representations, we apply multiple multi-level dilated residual (Dilated-Res) convolutional blocks to generate representations of semantic units with different lengths. Each Dilated-Res block, as shown in Figure \ref{fig:2}(b), is composed of two parallel modules that are referred to as the multi-level dilated convolutional module and the residual module. 

We introduce a dilated convolutional layer (DCNN) to learn the high-level semantic representations of the input texts. The concept of dilated convolution has become popular in semantic segmentation in computer vision \cite{Wang2018UnderstandingCF} and audio signal modeling \cite{vandenoord16_ssw}, and it has been applied to natural language processing tasks such as neural machine translation \cite{https://doi.org/10.48550/arxiv.1610.10099} and text classification \cite{lin-etal-2018-semantic-unit}. The main concept of DCNN expands the kernel by inserting holes between its consecutive elements in the filters, which aggregates multi-scale contextual information, such as words, phrases, and sentences. Inspired by \citet{lin-etal-2018-semantic-unit}, we use a three-level DCNN with different dilation rates to generate high-level semantic representations of the input texts. We input the word embeddings $E$ of the medical records into the 1-dimensional convolution operator with kernel size $K$ and dilation rates $[r_1, r_2, ..., r_m]$, where $m$ is the number of layers of 1-dimensional convolutions. The dilated convolutional procedure can be formalized:
\begin{equation}
    %\begin{gathered}
        H^i = (E\ast_{r}f)(s) = \sum_{j=0}^{K-1} f(i) \cdot E_{s-r\cdot j} \\
        %... \\
        %H^m = (E\ast_{r}f)(s) = \sum_{i=0}^{K-1} f(i) \cdot E_{s-r\cdot i},
    %\end{gathered}
\end{equation}
where $H^{i}$ is the output channel for layer $i=1,..., m$, $\ast$ denotes the convolution operation, $r$ is the dilation rate, $s$ is the element of the input sequence, $K$ is the kernel size, and $s-r\cdot j$ refers to past time steps. We force the length of the output of the $m$-layer 1-dimensional DCNN to be the same as the input $E$, to keep the sequence length unchanged after the convolution, that is, $H^{m} \in \mathbb{R}^{n \times {d_e}}$. To achieve this, we set a padding size $p = \frac{r(K-1)}{2}$ with a stride of $1$.  

In addition to the multi-level dilated convolutional module, we also simultaneously transform input embedding $E$ and add it to $H^{m}$ as in the residual network \cite{7780459}, which reduces the gradient vanishing issue in the deep encoder structure. We use a 1-dimensional convolutional layer with dilation rate $r_1$ to transform the input embedding $E$ into $\tilde{H}$. Then we add $\tilde{H}$ with $H^{m}$, the output from the multi-level dilated convolutional module, to form:
\begin{equation}
    D = \sigma(H^{m} + \tilde{H}),
\end{equation}
where  $D \in \mathbb{R}^{n \times {d_e}}$ represents the final clinical text, and $\sigma(\cdot)$ denotes an activation function.

\subsection{Label Encoder}
\subsubsection{Label Co-occurrence Encoding} % Via GCN}
The co-occurrence of disease codes in clinical texts is often observed when certain diseases are concurrent or have a causal relationship with each other. This means that the codes representing these related diseases tend to appear together in clinical text data. In order to capture the co-occurrence between disease codes in clinical texts, we construct a code co-occurrence graph. This is built using the code co-occurrence matrix, which serves as the adjacency matrix for the graph. To generate the code co-occurrence adjacency matrix, we model the correlation dependency between labels in terms of conditional probabilities. Specifically, we calculate the probability $P(L_j\,|\,L_i)$, which denotes the probability of occurrence of label $L_j$ when label $L_i$ appears. To facilitate graph construction, we binarize the correlation probability $P$. This entails converting the probability values into binary values, indicating whether a correlation exists between two labels. The operation can be written as:
\begin{equation}
    A_{ij} = 
    \begin{cases}
        0, \text{if } P < \lambda \\
        1, \text{if } P \ge \lambda, 
    \end{cases}
\end{equation}
where $A$ is the binary correlation matrix, and $\lambda$ is the hyper-parameter threshold to filter the noise edges. In our experiment, $\lambda = 1$, which means that the two labels in each pair will always appear together. With $\lambda = 1$, we obtain 39,166 ICD pairs (which are also the number of edges in the graph). Decreasing the probability rate causes an exponential growth of the number of edges, which greatly increases the complexity of the graph. We hypothesized that the performance of the model would not benefit from a denser graph, since a larger number of edges would result in overfitting. This hypothesis can be explored in the future. 

We employ a two-layer Graph Convolutional Network (GCN) to incorporate the co-occurrence relationships among labels. Specifically, we use the ICD full descriptors to generate a feature vector for each code. To calculate the feature vector for a specific code, we start by obtaining the word embeddings for each word in its descriptors, and then average the word embeddings to obtain a consolidated representation for the code:
\begin{equation}
     v_{i} = \frac{1}{Z}\sum_{j=1}^{Z} w_{j}, i=1, 2, ..., L,
\end{equation}
where $v_{i} \in \mathbb{R}^{d_{e}}$, $Z$ is the number of words in its descriptor, and $L$, the number of codes. The code vector set can be represented as $V = \{v_1,v_2,...,v_L\}$. In our graph structure, each node represents an ICD code, and the edges between nodes represent code co-occurrence relationships. In each layer of the GCN, the node features are aggregated based on these edge types to generate new label features for the subsequent layer:
\begin{equation}
    h^{l+1} = \sigma(A \cdot h^{l} \cdot W^{l}),
\end{equation}
where $h^{l}$ and $h^{l+1} \in \mathbb{R}^{L \times d_e}$ indicate the node representation of the $l^{th}$ and $(l+1)^{th}$ layers, $h^0 = V$, $A$ is the adjacency matrix of the label co-occurrence, $W$ is the layer-specific weight matrix and $\sigma(\cdot)$ denotes an activation function. We denote the last layer representation as $H_\textit{label} \in \mathbb{R}^{L \times d_e}$, which captures the code co-occurrence correlations.
 
\subsubsection{Dynamic Auxiliary Knowledge Mask} 
%Inspired by \citet{wang-etal-2022-kenmesh}, we dynamically generate a unique mask for each clinical note by integrating auxiliary knowledge in the EHR system. We use the mask to reduce the number of candidate ICD codes, as ICD codes have a wide range of occurrence frequencies. %, making them abundant and unevenly distributed. 
%Consequently, each ICD code has significantly more negative examples than positive ones. In order to improve the classifier's performance, a dynamic mask that is selected for each summary helps down-sample the negative examples and focus the classifier on candidate labels. 
ICD codes have a wide range of occurrence frequencies. Consequently, each ICD code has significantly more negative examples than positive ones. Inspired by \citet{wang-etal-2022-kenmesh}, to improve the classifier's performance, we dynamically generate a unique mask for each clinical note by integrating auxiliary knowledge in the EHR system. The dynamic mask that is selected for each summary helps down-sample the negative examples and focus the classifier on candidate labels. 
%Secondly, the original attention mechanism \cite{Bahdanau2015NeuralMT} poses a problem as it lacks specificity by focusing on identifying relevant information for all predicted labels, which can lead to irrelevant or non-pertinent information being identified. By using a masked label-wise attention approach, the classifier can find relevant information for each label within the ICD mask, which helps improve the pertinence and accuracy of the classification.

To generate the auxiliary knowledge masks, we consider three external knowledge sources: diagnosis-related group (DRG) codes, current procedural terminology (CPT) codes, and medications prescribed to patients. DRG codes are used to facilitate inpatient billing and reimbursement, and they categorize patients by their ICD codes and the cost associated with treatments. DRG codes are divided into medical DRGs (which don’t reflect operating room procedures) and surgical DRGs. CPT codes are used to describe clinical procedures and services in healthcare. They provide a standardized way of documenting and billing for medical services. Such code terminologies play a crucial role in improving ICD code predictions. Prescribed drugs also appear to be highly informative in predicting ICD codes, since they are often the final step of the episode of care. As patients near the end of their treatment or care, the prescribed medications play a crucial role in managing their conditions. Consequently, these medications serve as strong indicators or signals of the underlying health conditions or diagnoses, making them valuable in predicting the corresponding ICD codes.  %DRG codes are used by hospitals and healthcare providers to classify patients into groups based on their diagnosis, treatment, and length of stay. These codes are used for reimbursement purposes, and they help determine the amount of money that healthcare providers are paid for their services. CPT codes are used to describe medical procedures and services provided by healthcare providers. They provide a standardized way of documenting and billing for medical services. CPT codes are used by insurance companies to determine reimbursement rates for healthcare providers. We expect that DRG codes, CPT codes and prescribed drugs provide standardized ways of documenting and tracking medical care, and they play an important role in healthcare management and billing, which is crucial and related to ICD coding. 
We build an auxiliary knowledge-label co-occurrence matrix using conditional probabilities, i.e., $P(L_{i}\,|\,M_{j})$, which denote the probabilities of occurrence of label $L_{i}$ when auxiliary knowledge $M_{j}$ appears.   
\begin{equation} \label{eq:6}
    P(L_{i}\,|\,M_{j}) = \frac{C_{L_{i}\cap M_{j}}}{C_{M_{j}}},
\end{equation}
where $C_{L_{i}\cap M_{j}}$ denotes the number of co-occurrences of $L_{i}$ and $M_{j}$, and $C_{M_{j}}$ is the number of occurrences of $M_{j}$ in the training set. To avoid the noise of rare co-occurrences, a threshold $\tau$ filters noisy correlations.
$\tilde{M}_{j}$ denotes the selected ICD set for auxiliary knowledge $j$. 
\begin{equation} \label{eq:7}
    \tilde{M}_{j} = \{L_{k} \vert P(L_{k} \vert M_{j}) > \tau, \;
    k = 1, ..., L\}.
\end{equation}
We then join the ICD codes generated from the auxiliary knowledge co-occurrences for the DRG codes, CPT codes and prescribed drugs to form the final ICD mask set $T$:
\begin{equation} \label{eq:10}
    T = \tilde{M}_\textit{DRG} \cup \tilde{M}_\textit{CPT} \cup \tilde{M}_\textit{drug}.
\end{equation}
Then we assign a value to each label in $\mathcal{Y}$ to form $T_{\textit{vec}} \in [0,1]^{\mathcal{Y}}$. We assign 1 if the label appears in $T$, and 0 otherwise. The label order of $T_{\textit{vec}}$ is the same as $H_{\textit{label}}$. We then apply the mask to the label co-occurrence representation $H_{\textit{label}}$ to form:
\begin{equation}
    H_\textit{masked} = H_{\textit{label}} \odot T_{\textit{vec}},
\end{equation}
where $H_\textit{masked} \in \mathbb{R}^{L \times d_e}$ indicates the masked code representation. 

\subsection{Label-wise Attention Layer}
After encoding the clinical notes and their associated ICD codes, we obtain a clinical text representation denoted as $D$ and a masked code representation denoted as $H_\textit{masked}$. As we aim to assign multiple codes to each clinical note and recognize that different codes may be relevant to different sections of the document, we employ a code-wise attention mechanism. This mechanism allows the model to learn the relevant document representations specific to each code. To generate the code-wise attention vector, we use a matrix-vector product:
\begin{equation}
    \alpha = \textrm{softmax}(D \cdot H_{\textit{masked}})
\end{equation}
Finally, we leverage the document representation $D$ and corresponding code-wise attention vector $\alpha$ to generate the code-aware document representation:
\begin{equation}
    C = \alpha \cdot D,
\end{equation}
where $C \in \mathbb{R}^{L \times d_e}$. 
\subsection{Classifier}
To perform classification, we compute the probability for each code by using a fully connected layer followed by a \textit{sigmoid} transformation:
\begin{equation}
    \hat{y} = \textrm{sigmoid}(W \cdot C), 
\end{equation}
where $W \in \mathbb{R}^{d_e \times 1}$ indicates the weight matrix. Our model is trained using the multi-label binary cross-entropy loss:
\begin{equation}
    L =\sum_{i=1}^{L}[-y_{i} \cdot \log(\hat{y_{i}}) -  (1-y_{i}) \cdot \log(1 - \hat{y_{i}})],
\end{equation}
where $y_i$ is the ground truth of code $i$. 
\section{Experiments}
\subsection{Dataset and pre-processing}
We use the MIMIC-III dataset \citelanguageresource{Johnson2016MIMICIIIAF}, which is the largest publicly available clinical dataset for text, and  comprises hospital records associated with over 40,000 patients. We focus on the discharge summaries that are human expert-labeled with a set of ICD-9 codes. We follow the experimental setting of \citet{mullenbach-etal-2018-explainable} to form MIMIC-III-full and MIMIC-III-top 50. To pre-process the clinical notes, we first remove all deidentified information and replace punctuation and atypical alphanumerical character combinations (e.g., 3a, 4kg) with white space. We then transform every token into its lowercase. The maximum length of a token sequence is 4,000 and any that exceed this length is truncated.  

\subsection{Evaluation Metrics and Implementation Details}
Following previous work \cite{mullenbach-etal-2018-explainable}, we evaluate our method using both macro and micro F1 and AUC metrics, as well as precision at $K$ ($P@K$) that indicates the proportion of the correctly predicted labels in the top-$K$ predictions. 

We implement our model in PyTorch \cite{NEURIPS2019_9015} on a single NVIDIA A100 40G GPU. The training process completes in 5 hours, while inference requires only 30 minutes (about 0.5s per note). This performance underscores the computational efficiency of our method, indicating it is not only effective in handling complex tasks but also practical in terms of computational resources and time. We use word2vec \cite{NIPS2013_9aa42b31}  to pre-train the word embeddings of dimension 100 on the pre-processed  MIMIC-III texts. We use a three-level dilated convolution with dilation rate [1, 2, 4], and the filter size of the convolution is 9. We use the Adam optimizer and early stopping strategies. The learning rate is initialized to 0.0001, and the decay rate is 0.9 in every epoch. The gradient clip is applied to the maximum norm of 5. The batch size is 32. Table \ref{table:1} shows our detailed hyper-parameter settings. We evaluate with 5 different random seeds for the model and report the average test results. Our code is available at \urlstyle{same}\url{https://github.com/xdwang0726/MIMIC-ICD-Classification}.
\begin{table}[t] 
\begin{center}
\resizebox{\columnwidth}{!}{
\begin{tabular}{ l c }
      \hline
      Hyper-parameters & Values\\
      \hline
      embedding size & \textbf{100}, 200\\
   %   \hline
      filter size & 3, 5, \textbf{9}\\
   %   \hline
      prediction threshold & 0.0005\\
   %   \hline
      dropout & \textbf{0.2}, 0.5 \\
   %   \hline
      dilation rate & \textbf{[1, 2, 4]}, [2, 5, 9]\\
   %   \hline
      learning rate &  \textbf{0.0001}, 0.0003, 0.0005\\
  %    \hline
      batch size & 8, 16, \textbf{32}\\
      \hline
\end{tabular}}
\caption{Hyper-parameter settings. Bold: the optimal values.} \label{table:1}
\end{center}
\end{table}

\section{Results and Discussions}

\begin{table*}[t]
\resizebox{\textwidth}{!}{
\begin{tabular}{c cccccc  ccccc}
\hline
\multirow{3}{*}{Models} & \multicolumn{6}{c}{MIMIC-III-full}                                         & \multicolumn{5}{c}{MIMIC-III-top 50}                                    \\ \cline{2-12} 
                        & \multicolumn{2}{c}{AUC} & \multicolumn{2}{c}{F1} & \multicolumn{2}{c}{P@K} & \multicolumn{2}{c}{AUC} & \multicolumn{2}{c}{F1} & \multirow{2}{*}{P@5} \\ \cline{2-11}
                        & Macro      & Micro      & Macro      & Micro     & P@8        & P@15       & Macro      & Micro      & Macro      & Micro     &                      \\ \hline
CAML \cite{mullenbach-etal-2018-explainable}                   & 0.895      & 0.986      & 0.088      & 0.539     & 0.709      & 0.561      & 0.875      & 0.909      & 0.532      & 0.614     & 0.609                \\
DR-CAML \cite{mullenbach-etal-2018-explainable}                & 0.897      & 0.985      & 0.086      & 0.529     & 0.690      & 0.548      & 0.884      & 0.916      & 0.576      & 0.633     & 0.618                \\
MultiResCNN \cite{Li2020ICDCF}             & 0.910      & 0.986      & 0.085      & 0.552     & 0.734      & 0.584      & 0.899      & 0.928      & 0.606      & 0.670     & 0.641                \\
LAAT \cite{ijcai2020-461}                   & 0.919      & 0.988      & 0.099      & 0.575     & 0.738      & 0.591      & 0.925      & 0.946      & 0.666      & 0.715     & 0.675                \\
Joint-LAAT \cite{ijcai2020-461}             & 0.921      & 0.988      & 0.107      & 0.575     & 0.735      & 0.590      & 0.925      & 0.946      & 0.661      & 0.716     & 0.671                \\
EffectiveCAN  \cite{liu-etal-2021-effective}          & 0.915      & 0.988      & 0.106      & 0.589     & 0.758      & 0.606      & 0.915      & 0.938      & 0.644      & 0.702     & 0.656                \\
MSMN \cite{yuan-etal-2022-code}                   & \textbf{0.950}      & 0.992      & 0.103      & 0.584     & 0.752      & 0.599      & 0.928      & 0.947      & 0.683      & 0.725     & 0.680                \\
KEPTLongformer \cite{yang-etal-2022-knowledge-injected}         & -          & -          & \textbf{0.118}      & 0.599     & 0.771      & 0.615      & 0.926      & 0.947      & 0.689      & 0.728     & 0.672                \\ \hline
\multirow{2}{*}{Ours}                    & 0.948      & \textbf{0.994}      & 0.112      & \textbf{0.605}     & \textbf{0.784}      & \textbf{0.637}           &     \textbf{0.928}       &    \textbf{0.950}        &    \textbf{0.692}       &    \textbf{0.734}     & \textbf{0.683} \\
~ & $\pm$ 0.022 & $\pm$ 0.013 & $\pm$ 0.027 & $\pm$ 0.021 & $\pm$ 0.022  & $\pm$ 0.011 & $\pm$ 0.014 & $\pm$ 0.018 & $\pm$ 0.016 & $\pm$ 0.012 & $\pm$ 0.023 \\
\hline                   
\end{tabular}}
\caption{Comparison to previous methods across three main evaluation metrics MIMIC-III dataset. We report the mean $\pm$ standard deviation of each result. Bold: best scores in each column.} \label{table:2}
\end{table*}

\begin{table}[t]
\resizebox{\columnwidth}{!}{
\begin{tabular}{ccccc}
\hline
\multirow{2}{*}{Methods} & \multicolumn{2}{c}{AUC} & \multicolumn{2}{c}{P@K} \\ \cline{2-5} 
                        & Macro      & Micro      & P@8        & P@15       \\ \hline
Full Model               & \textbf{0.948}      & \textbf{0.994}      & \textbf{0.784}      & \textbf{0.637}      \\ \hline
embedded w/ Longformer    & 0.918      & 0.987      & 0.751       & 0.592    \\
embedded w/ BioWordVec   & 0.923      & 0.989      & 0.765      & 0.609      \\
w/o label feature       & 0.904      & 0.986      & 0.736      & 0.583      \\
w/o masked attention     & 0.912           &   0.986         &    0.756        &   0.592         \\ \hline
\end{tabular}}
\caption{Ablation experiment results. Bold: the optimal values.} \label{table:3}
\end{table}

To evaluate the effectiveness of our proposed model, we compare with existing state-of-the-art %(SOTA) 
methods, which are given in Table \ref{table:2}. Each row represents the evaluation metrics for a specific method. The best score for each metric is highlighted. According to the reported results, our model demonstrates superior performance across the majority of evaluation metrics, with the exceptions of Macro-AUC and Macro-F1 on the MIMIC-III-full dataset. Under the Top-50 codes setting, our model performs better than the KEPTLongformer on all metrics and achieves state-of-the-art scores. These %outstanding 
results confirm the effectiveness of leveraging auxiliary knowledge and label co-occurrence relations. 

\subsection{Ablation Studies}
We aim to investigate the influence of various modules of our model, and we seek to understand how these modules contribute to the performance of the model in terms of both effectiveness and robustness. In order to conduct a fair comparison and isolate the effects of specific modules, we systematically remove certain modules from our model. Specifically, we conduct controlled experiments with three different settings: (a) examining the influence of different embedding methods by replacing built from scratch embeddings with pre-trained %language model, 
contextual embeddings, i.e., Clinical-Longformer \cite{li2023comparative} and pre-trained biomedical context-free embeddings, i.e., BioWordVec \cite{Zhang2019BioWordVecIB}; (b) replacing the co-occurrence graph learning with a fully connected layer; (c) removing the auxiliary knowledge mask from our model. The experimental results are shown in Table \ref{table:3}. 

\textbf{Effectiveness of Embedding Methods}
As shown in Table \ref{table:3}, using pre-trained context-free word embeddings (BioWordVec) and pre-trained contextual embeddings (Clinical-Longformer) have negative impacts on the performance. This observation shows that although the use of pre-trained word embeddings has shown impressive performance across a wide range of natural language processing tasks, their performance on clinical datasets can sometimes be suboptimal. Understanding why this can occur will require further study. %The reason for this 

\textbf{Effectiveness of Learning Label Representations} Table \ref{table:3} shows the positive contribution of label representations learned by GCN. By using GCN, our model gains the ability to capture and leverage the relationships and dependencies between labels, leading to improvements in performance. This indicates that the incorporation of label co-occurrence information in a %through 
GCN enables the model to learn from the collective behaviour of labels, facilitating a more comprehensive understanding of the underlying label relationships.

\textbf{Effectiveness of Involving Auxiliary Knowledge Mask} We have three types of auxiliary knowledge involved to build the mask: DRG codes, CPT codes, and drug prescriptions. As reported in Table \ref{table:3},  performance drops when removing the auxiliary knowledge mask, suggesting that the auxiliary knowledge mask plays a crucial role in guiding the model's attention towards relevant information and aiding in the classification process. This result provides further evidence supporting the premise that the auxiliary knowledge mask effectively leverages external knowledge to mitigate the challenges posed by an extensive pool of potential ICD codes. By incorporating external knowledge through the auxiliary knowledge mask, the model gains the ability to narrow down and focus on relevant labels, thereby enhancing its efficiency and accuracy in the final prediction. To select the proper mask for each clinical note, one hyper-parameter is used: threshold $\tau$ of auxiliary knowledge-label co-occurrence. With $\tau = 0.005$, 99.22\% of the gold-standard
ICD codes are guaranteed to be in the mask, and the average number of codes in the mask is 1460 which is about $\frac{1}{6}$ of the complete set of codes. 

\subsection{Case Studies}
\begin{figure}[t]
\includegraphics[width=\columnwidth]{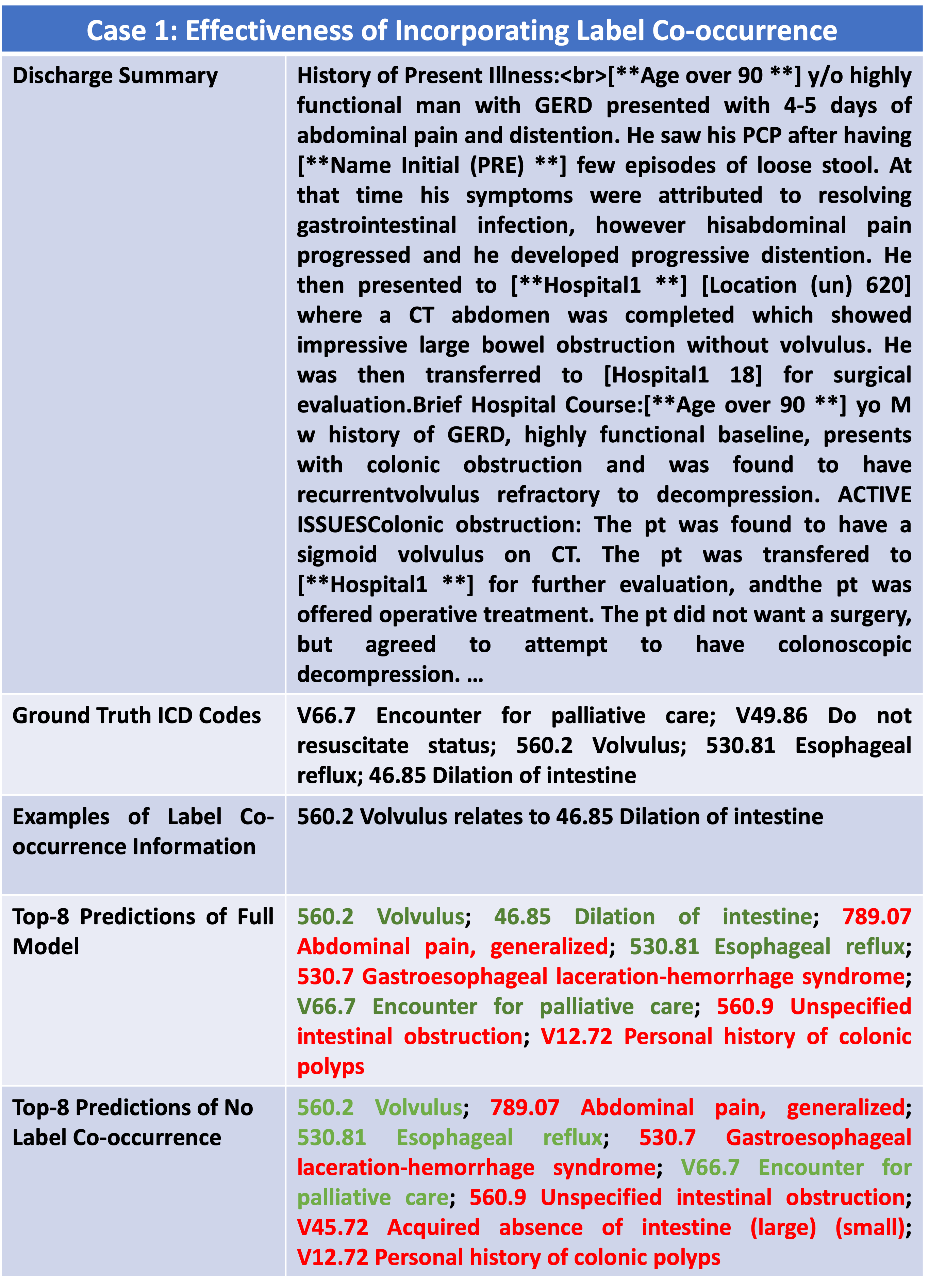}
\caption{Case study on the effectiveness of incorporating label co-occurrence. Correctly predicted labels are marked in green and the incorrect ones are marked in red.}
\label{fig:3}
\end{figure}

\begin{figure}[t]
\vspace{.5mm}
\includegraphics[width=\columnwidth]{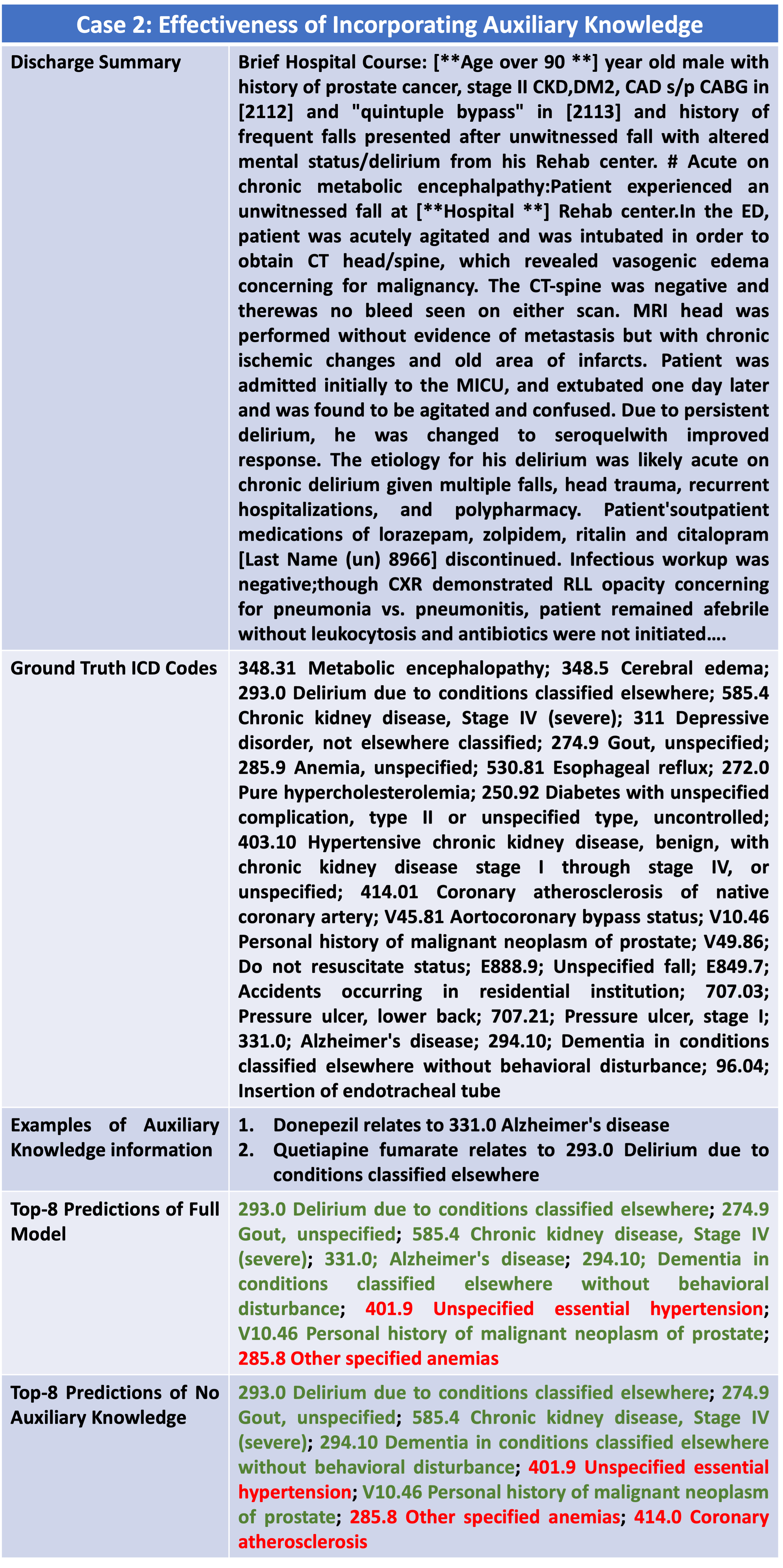}
\caption{Case study on the effectiveness of incorporating auxiliary knowledge. Correctly predicted labels are marked in green and the incorrect ones are marked in red.}
\label{fig:4}
\end{figure}

We conduct case studies to qualitatively understand the effects of incorporating the label co-occurrence (as shown in Figure \ref{fig:3}) and the auxiliary knowledge (as shown in Figure \ref{fig:4}). %We compare the full model with model without using label feature and model without using masked attention on the predictions of two patient records. 
For each patient, we show the discharge summary, ground truth ICD codes, label co-occurrence information / auxiliary knowledge information as well as the top-8 predicted ICD codes of the full model and ablated models. In Case 1, the ground truth ICD codes include ``46.85 Dilation of intestine'', a diagnosis not explicitly mentioned in the discharge summary. The observed label co-occurrence between ``560.2 Volvulus'' and ``46.85 Dilation of intestine'' serves as a robust indicator, effectively suggesting the presence of the ``46.85 Dilation of intestine'' diagnosis in the patient. Without the label co-occurrence signals, the ablated model makes a wrong prediction ``789.07 Abdominal pain, generalized'' that ignores the latent label information. In Case 2, the patient has been diagnosed with ``331.0 Alzheimer's disease'' with less explicit information in the discharge summary. Notably, the presence of ``Donepezil'' in the drug prescription, an element of the auxiliary knowledge, indicates that the patient is most likely to have Alzheimer’s disease. The ablated model, lacking the auxiliary knowledge, mistakenly predicts ``285.8 Other specified anemias''. Case 1 and Case 2 exemplify the advantages of incorporating label co-occurrence and auxiliary knowledge, respectively.

\section{Conclusion}
In this paper, we propose a novel auxiliary knowledge-induced medical code labelling framework which uses multiple multi-level dilated residual blocks and jointly exploits label co-occurrence and auxiliary knowledge. Specifically,  incorporating label co-occurrence relations and external knowledge through the auxiliary knowledge mask serves as a valuable mechanism for addressing the inherent complexity and size of the label space, ultimately leading to improved performance and more effective utilization of the model's resources. Moreover, to deal with the length of the clinical texts, the multi-level dilated residual block helps capture and understand long dependencies. Experimental results demonstrate that our proposed model outperforms the baseline models.  We are interested in integrating more external knowledge in the future, such as the Unified Medical Language System (UMLS), to seek further improvements. 

\section{Limitations}
Our work is limited to evaluate the MIMIC-III-full and MIMIC-III-top 50, which are mostly focused on  common diseases (i.e., the most frequent ICD codes). It is not possible to define rare diseases simply from the distribution of ICD codes in the dataset since rare ICD codes do not necessarily indicate the presence of rare diseases exclusively. This limits evaluation on diseases that are rare {\em a priori}. A list of rare diseases proposed by domain experts for more specific medical tasks would be helpful to explore more focused use cases.

Our auxiliary knowledge masks are limited by external knowledge including DRG codes, CPT codes, and drug prescriptions. Other knowledge sources, including disease-symptom, disease-lab relations, for example, could potentially be useful for the auto ICD coding task.

\section*{Acknowledgements}
We would like to thank all reviewers for their comments, which helped improve this paper considerably. Computational resources used in preparing this research were provided, in part, by Compute Ontario\footnote{\urlstyle{same}\url{https://www.computeontario.ca}}, Digital Research Alliance of Canada\footnote{\urlstyle{same}\url{https://ccdb.alliancecan.ca}}, the Province of Ontario, the Government of Canada through CIFAR, and companies sponsoring the Vector Institute\footnote{\urlstyle{same}\url{https://www.vectorinstitute.ai/partners}}. This research is partially funded by The Natural Sciences and Engineering Research Council of Canada (NSERC) through a Discovery Grant to R. E. Mercer. F. Rudzicz is supported by a CIFAR Chair in AI.
%\nocite{*}
\section{Bibliographical References}\label{sec:reference}

\bibliographystyle{lrec-coling2024-natbib}
\bibliography{lrec-coling2024-example}

\begin{thebibliography}{1}
\expandafter\ifx\csname natexlab\endcsname\relax\def\natexlab#1{#1}\fi

\bibitem[{Johnson et~al.(2016)Johnson, Pollard, Shen, Lehman, Feng, Ghassemi, Moody, Szolovits, Anthony~Celi, and Mark}]{Johnson2016MIMICIIIAF}
Johnson, Alistair EW and Pollard, Tom J and Shen, Lu and Lehman, Li-wei H and Feng, Mengling and Ghassemi, Mohammad and Moody, Benjamin and Szolovits, Peter and Anthony Celi, Leo and Mark, Roger G. 2016.
\newblock \emph{MIMIC-III, a freely accessible critical care database}.
\newblock Nature Publishing Group.

\end{thebibliography}


\begin{thebibliography}{34}
\expandafter\ifx\csname natexlab\endcsname\relax\def\natexlab#1{#1}\fi

\bibitem[{Bai and Vucetic(2019)}]{Bai2019ImprovingMC}
Tian Bai and Slobodan Vucetic. 2019.
\newblock Improving medical code prediction from clinical text via incorporating online knowledge sources.
\newblock \emph{The World Wide Web Conference}, pages 72--82.

\bibitem[{Baumel et~al.(2018)Baumel, Nassour-Kassis, Elhadad, and Elhadad}]{Baumel2018MultiLabelCO}
Tal Baumel, Jumana Nassour-Kassis, Michael Elhadad, and No{\'e}mie Elhadad. 2018.
\newblock Multi-label classification of patient notes a case study on {ICD} code assignment.
\newblock In \emph{The Workshops of the Thirty-Second {AAAI} Conference on Artificial Intelligence}, pages 409--416.

\bibitem[{Cao et~al.(2020)Cao, Chen, Liu, Zhao, Liu, and Chong}]{Cao2020HyperCoreHA}
Pengfei Cao, Yubo Chen, Kang Liu, Jun Zhao, Shengping Liu, and Weifeng Chong. 2020.
\newblock Hypercore: Hyperbolic and co-graph representation for automatic icd coding.
\newblock In \emph{Proceedings of the 58th Annual Meeting of the Association for Computational Linguistics (Volume 1: Long Papers)}, pages 3105--3114.

\bibitem[{Crammer et~al.(2007)Crammer, Dredze, Ganchev, Talukdar, and Carroll}]{10.5555/1572392.1572416}
Koby Crammer, Mark Dredze, Kuzman Ganchev, Partha~Pratim Talukdar, and Steven Carroll. 2007.
\newblock Automatic code assignment to medical text.
\newblock In \emph{Proceedings of the Workshop on BioNLP 2007: Biological, Translational, and Clinical Language Processing}, pages 129--136.

\bibitem[{de~Lima et~al.(1998)de~Lima, Laender, and Ribeiro-Neto}]{10.1145/288627.288649}
Luciano R.~S. de~Lima, Alberto H.~F. Laender, and Berthier~A. Ribeiro-Neto. 1998.
\newblock A hierarchical approach to the automatic categorization of medical documents.
\newblock In \emph{Proceedings of the Seventh International Conference on Information and Knowledge Management}, CIKM '98, pages 132--139.

\bibitem[{Farkas and Szarvas(2008)}]{Farkas2008AutomaticCO}
Rich{\'a}rd Farkas and Gy{\"o}rgy Szarvas. 2008.
\newblock Automatic construction of rule-based {ICD-9-CM} coding systems.
\newblock \emph{BMC Bioinformatics}, 9:S10 -- S10.

\bibitem[{He et~al.(2016)He, Zhang, Ren, and Sun}]{7780459}
Kaiming He, Xiangyu Zhang, Shaoqing Ren, and Jian Sun. 2016.
\newblock \href {https://doi.org/10.1109/CVPR.2016.90} {Deep residual learning for image recognition}.
\newblock In \emph{2016 IEEE Conference on Computer Vision and Pattern Recognition (CVPR)}, pages 770--778.

\bibitem[{Ji et~al.(2020)Ji, Cambria, and Marttinen}]{ji-etal-2020-dilated}
Shaoxiong Ji, Erik Cambria, and Pekka Marttinen. 2020.
\newblock Dilated convolutional attention network for medical code assignment from clinical text.
\newblock In \emph{Proceedings of the 3rd Clinical Natural Language Processing Workshop}, pages 73--78.

\bibitem[{Kalchbrenner et~al.(2016)Kalchbrenner, Espeholt, Simonyan, Oord, Graves, and Kavukcuoglu}]{https://doi.org/10.48550/arxiv.1610.10099}
Nal Kalchbrenner, Lasse Espeholt, Karen Simonyan, Aaron van~den Oord, Alex Graves, and Koray Kavukcuoglu. 2016.
\newblock Neural machine translation in linear time.
\newblock \emph{arXiv}, 1610.10099v2.

\bibitem[{Kipf and Welling(2017)}]{Kipf:2016tc}
Thomas~N. Kipf and Max Welling. 2017.
\newblock Semi-supervised classification with graph convolutional networks.
\newblock In \emph{Proceedings of the 5th International Conference on Learning Representations}.

\bibitem[{Larkey and Croft(1996)}]{Larkey1996CombiningCI}
Leah~S. Larkey and W.~Bruce Croft. 1996.
\newblock Combining classifiers in text categorization.
\newblock In \emph{Proceedings of the 19th Annual International ACM SIGIR Conference on Research and Development in Information Retrieval}, pages 289--297.

\bibitem[{Li and Yu(2020)}]{Li2020ICDCF}
Fei Li and Hong Yu. 2020.
\newblock {ICD} coding from clinical text using multi-filter residual convolutional neural network.
\newblock In \emph{Proceedings of the Thirty-Fourth AAAI Conference on Artificial Intelligence}, pages 8180--8187.

\bibitem[{Li et~al.(2023)Li, Wehbe, Ahmad, Wang, and Luo}]{li2023comparative}
Yikuan Li, Ramsey~M Wehbe, Faraz~S Ahmad, Hanyin Wang, and Yuan Luo. 2023.
\newblock A comparative study of pretrained language models for long clinical text.
\newblock \emph{Journal of the American Medical Informatics Association}, 30(2):340--347.

\bibitem[{Lin et~al.(2018)Lin, Su, Yang, Ma, and Sun}]{lin-etal-2018-semantic-unit}
Junyang Lin, Qi~Su, Pengcheng Yang, Shuming Ma, and Xu~Sun. 2018.
\newblock Semantic-unit-based dilated convolution for multi-label text classification.
\newblock In \emph{Proceedings of the 2018 Conference on Empirical Methods in Natural Language Processing}, pages 4554--4564.

\bibitem[{Lita et~al.(2008)Lita, Yu, Niculescu, and Bi}]{lita-etal-2008-large}
Lucian~Vlad Lita, Shipeng Yu, Stefan Niculescu, and Jinbo Bi. 2008.
\newblock Large scale diagnostic code classification for medical patient records.
\newblock In \emph{Proceedings of the Third International Joint Conference on Natural Language Processing: Volume-{II}}, pages 877--882.

\bibitem[{Liu et~al.(2017)Liu, Chang, Wu, and Yang}]{10.1145/3077136.3080834}
Jingzhou Liu, Wei-Cheng Chang, Yuexin Wu, and Yiming Yang. 2017.
\newblock Deep learning for extreme multi-label text classification.
\newblock In \emph{Proceedings of the 40th International {ACM} {SIGIR} Conference on Research and Development in Information Retrieval}, pages 115--124.

\bibitem[{Liu et~al.(2021)Liu, Cheng, Klopfer, Gormley, and Schaaf}]{liu-etal-2021-effective}
Yang Liu, Hua Cheng, Russell Klopfer, Matthew~R. Gormley, and Thomas Schaaf. 2021.
\newblock Effective convolutional attention network for multi-label clinical document classification.
\newblock In \emph{Proceedings of the 2021 Conference on Empirical Methods in Natural Language Processing}, pages 5941--5953.

\bibitem[{Mikolov et~al.(2013{\natexlab{a}})Mikolov, Sutskever, Chen, Corrado, and Dean}]{10.5555/2999792.2999959}
Tomas Mikolov, Ilya Sutskever, Kai Chen, Greg Corrado, and Jeffrey Dean. 2013{\natexlab{a}}.
\newblock Distributed representations of words and phrases and their compositionality.
\newblock In \emph{Proceedings of the 26th International Conference on Neural Information Processing Systems - Volume 2}, NIPS'13, pages 3111--3119.

\bibitem[{Mikolov et~al.(2013{\natexlab{b}})Mikolov, Sutskever, Chen, Corrado, and Dean}]{NIPS2013_9aa42b31}
Tomas Mikolov, Ilya Sutskever, Kai Chen, Greg~S Corrado, and Jeff Dean. 2013{\natexlab{b}}.
\newblock \href {https://proceedings.neurips.cc/paper_files/paper/2013/file/9aa42b31882ec039965f3c4923ce901b-Paper.pdf} {Distributed representations of words and phrases and their compositionality}.
\newblock In \emph{Advances in Neural Information Processing Systems}, volume~26.

\bibitem[{Mullenbach et~al.(2018)Mullenbach, Wiegreffe, Duke, Sun, and Eisenstein}]{mullenbach-etal-2018-explainable}
James Mullenbach, Sarah Wiegreffe, Jon Duke, Jimeng Sun, and Jacob Eisenstein. 2018.
\newblock Explainable prediction of medical codes from clinical text.
\newblock In \emph{Proceedings of the 2018 Conference of the North {A}merican Chapter of the Association for Computational Linguistics: Human Language Technologies, Volume 1 (Long Papers)}, pages 1101--1111.

\bibitem[{O'Malley et~al.(2005)O'Malley, Cook, Price, Wildes, Hurdle, and Ashton}]{b0be0de6987549188fdfae53aa2a6554}
{Kimberly J.} O'Malley, {Karon F.} Cook, {Matt D.} Price, {Kimberly Raiford} Wildes, {John F.} Hurdle, and {Carol M.} Ashton. 2005.
\newblock \href {https://doi.org/10.1111/j.1475-6773.2005.00444.x} {Measuring diagnoses: {ICD} code accuracy}.
\newblock \emph{Health Services Research}, 40(5 II):1620--1639.

\bibitem[{Paszke et~al.(2019)Paszke, Gross, Massa, Lerer, Bradbury, Chanan, Killeen, Lin, Gimelshein, Antiga, Desmaison, Kopf, Yang, DeVito, Raison, Tejani, Chilamkurthy, Steiner, Fang, Bai, and Chintala}]{NEURIPS2019_9015}
Adam Paszke, Sam Gross, Francisco Massa, Adam Lerer, James Bradbury, Gregory Chanan, Trevor Killeen, Zeming Lin, Natalia Gimelshein, Luca Antiga, Alban Desmaison, Andreas Kopf, Edward Yang, Zachary DeVito, Martin Raison, Alykhan Tejani, Sasank Chilamkurthy, Benoit Steiner, Lu~Fang, Junjie Bai, and Soumith Chintala. 2019.
\newblock {PyTorch}: An imperative style, high-performance deep learning library.
\newblock In \emph{Advances in Neural Information Processing Systems}, volume~32, pages 8024--8035.

\bibitem[{Rios and Kavuluru(2018)}]{rios-kavuluru-2018-shot}
Anthony Rios and Ramakanth Kavuluru. 2018.
\newblock Few-shot and zero-shot multi-label learning for structured label spaces.
\newblock In \emph{Proceedings of the 2018 Conference on Empirical Methods in Natural Language Processing}, pages 3132--3142.

\bibitem[{Shi et~al.(2017)Shi, Xie, Hu, Zhang, and Xing}]{Shi2017TowardsAI}
Haoran Shi, Pengtao Xie, Zhiting Hu, Ming Zhang, and Eric~P. Xing. 2017.
\newblock Towards automated {ICD} coding using deep learning.
\newblock \emph{ArXiv}, abs/1711.04075.

\bibitem[{{van den Oord} et~al.(2016){van den Oord}, Dieleman, Zen, Simonyan, Vinyals, Graves, Kalchbrenner, Senior, and Kavukcuoglu}]{vandenoord16_ssw}
Aäron {van den Oord}, Sander Dieleman, Heiga Zen, Karen Simonyan, Oriol Vinyals, Alex Graves, Nal Kalchbrenner, Andrew Senior, and Koray Kavukcuoglu. 2016.
\newblock {W}ave{N}et: A generative model for raw audio.
\newblock In \emph{Proceedings of the 9th ISCA Workshop on Speech Synthesis Workshop (SSW 9)}, page 125.

\bibitem[{Vu et~al.(2020)Vu, Nguyen, and Nguyen}]{ijcai2020-461}
Thanh Vu, Dat~Quoc Nguyen, and Anthony Nguyen. 2020.
\newblock A label attention model for {ICD} coding from clinical text.
\newblock In \emph{Proceedings of the Twenty-Ninth International Joint Conference on Artificial Intelligence, {IJCAI-20}}, pages 3335--3341.
\newblock Main track.

\bibitem[{Wang et~al.(2018)Wang, Chen, Yuan, Liu, Huang, Hou, and Cottrell}]{Wang2018UnderstandingCF}
Panqu Wang, Pengfei Chen, Ye~Yuan, Ding Liu, Zehua Huang, Xiaodi Hou, and G.~Cottrell. 2018.
\newblock Understanding convolution for semantic segmentation.
\newblock \emph{2018 IEEE Winter Conference on Applications of Computer Vision (WACV)}, pages 1451--1460.

\bibitem[{Wang et~al.(2022)Wang, Mercer, and Rudzicz}]{wang-etal-2022-kenmesh}
Xindi Wang, Robert Mercer, and Frank Rudzicz. 2022.
\newblock {K}en{M}e{SH}: Knowledge-enhanced end-to-end biomedical text labelling.
\newblock In \emph{Proceedings of the 60th Annual Meeting of the Association for Computational Linguistics (Volume 1: Long Papers)}, pages 2941--2951.

\bibitem[{Xie and Xing(2018)}]{xie-xing-2018-neural}
Pengtao Xie and Eric Xing. 2018.
\newblock A neural architecture for automated {ICD} coding.
\newblock In \emph{Proceedings of the 56th Annual Meeting of the Association for Computational Linguistics (Volume 1: Long Papers)}, pages 1066--1076.

\bibitem[{Xie et~al.(2019)Xie, Xiong, Yu, and Zhu}]{10.1145/3357384.3357897}
Xiancheng Xie, Yun Xiong, Philip~S. Yu, and Yangyong Zhu. 2019.
\newblock {EHR} coding with multi-scale feature attention and structured knowledge graph propagation.
\newblock In \emph{Proceedings of the 28th ACM International Conference on Information and Knowledge Management}, CIKM '19, pages 649--658.

\bibitem[{Yang et~al.(2022)Yang, Wang, Rawat, Mitra, and Yu}]{yang-etal-2022-knowledge-injected}
Zhichao Yang, Shufan Wang, Bhanu Pratap~Singh Rawat, Avijit Mitra, and Hong Yu. 2022.
\newblock \href {https://aclanthology.org/2022.findings-emnlp.127} {Knowledge injected prompt based fine-tuning for multi-label few-shot {ICD} coding}.
\newblock In \emph{Findings of the Association for Computational Linguistics: EMNLP 2022}, pages 1767--1781.

\bibitem[{You et~al.(2019)You, Zhang, Wang, Dai, Mamitsuka, and Zhu}]{10.5555/3454287.3454810}
Ronghui You, Zihan Zhang, Ziye Wang, Suyang Dai, Hiroshi Mamitsuka, and Shanfeng Zhu. 2019.
\newblock {AttentionXML}: Label tree-based attention-aware deep model for high-performance extreme multi-label text classification.
\newblock In \emph{Proceedings of the 33rd International Conference on Neural Information Processing Systems}, pages 5820--5830.

\bibitem[{Yuan et~al.(2022)Yuan, Tan, and Huang}]{yuan-etal-2022-code}
Zheng Yuan, Chuanqi Tan, and Songfang Huang. 2022.
\newblock \href {https://doi.org/10.18653/v1/2022.acl-short.91} {Code synonyms do matter: Multiple synonyms matching network for automatic {ICD} coding}.
\newblock In \emph{Proceedings of the 60th Annual Meeting of the Association for Computational Linguistics (Volume 2: Short Papers)}, pages 808--814.

\bibitem[{Zhang et~al.(2019)Zhang, Chen, Yang, Lin, and Lu}]{Zhang2019BioWordVecIB}
Yijia Zhang, Qingyu Chen, Zhihao Yang, Hongfei Lin, and Zhiyong Lu. 2019.
\newblock {BioWordVec, improving biomedical word embeddings with subword information and MeSH}.
\newblock \emph{Scientific Data}, 6.

\end{thebibliography}

\section{Language Resource References}
\label{lr:ref}
\bibliographystylelanguageresource{lrec-coling2024-natbib}
\bibliographylanguageresource{languageresource}

\end{document}